# Intuitive Robot Programming by Capturing Human Manufacturing Skills: A Framework for the Process of Glass Adhesive Application

Mihail Babcinschi[1], Francisco Cruz[1], Nicole Duarte[1], Silvia Santos[1], Samuel Alves[1] and Pedro Neto[1]

[1] Centre for Mechanical Engineering, Materials and Processes, Department of Mechanical Engineering, University of Coimbra, Portugal
pedro.neto@dem.uc.pt

**Abstract.** There is a great demand for the robotization of manufacturing processes featuring monotonous labor. Some manufacturing tasks requiring specific skills (welding, painting, etc.) suffer from a lack of workers. Robots have been used in these tasks, but their flexibility is limited since they are still difficult to program/re-program by non-experts, making them inaccessible to most companies. Robot offline programming (OLP) is reliable. However, generated paths directly from CAD/CAM do not include relevant parameters representing human skills such as robot end-effector orientations and velocities. This paper presents an intuitive robot programming system to capture human manufacturing skills and transform them into robot programs. Demonstrations from human skilled workers are recorded using a magnetic tracking system attached to the worker tools. Collected data include the orientations and velocity of the working paths. Positional data are extracted from CAD/CAM since its error when captured by the magnetic tracker, is significant. Paths poses are transformed in Cartesian space and validated in a simulation environment. Robot programs are generated and transferred to the real robot. Experiments on the process of glass adhesive application demonstrated the intuitiveness to use and effectiveness of the proposed framework in capturing human skills and transferring them to the robot.

**Keywords:** Robotics, Manufacturing Skills, Human-Robot Interfaces.

## 1    Introduction

In recent decades we have witnessed the robotization of various manufacturing processes requiring intensive and monotonous human labor. In addition, there is a lack of skilled labor to perform manufacturing tasks such as welding, painting, etc. In such a scenario, robotization appears as a solution for companies struggling to find skilled workers. The complexity of the process of robotization depends on the specific application and its integration into the industrial ecosystem. Normally, end-user companies contract a robot integrator to implement and integrate the robotic systems on the factory floor. The problem is that most of these end-user companies do not have the necessary human resources and skills to operate robots, especially when the production system is updated and robots need to be re-programed. This is an issue in the current context of mass customization featuring frequent updates in the production system and where robots are still difficult to program/re-program by non-experts. It is critical that robots can be interfaced and programmed in an intuitive way so that they can be operated by non-experts, making them accessible to companies.

In recent years research has led to enormous advancements toward the development of user-friendly robot interfaces and robots that collaborate and share the workspace with humans [1]. Nevertheless, they are still difficult to program by non-experts in robotics. Offline robot programming (OLP) demonstrated reliability in generating programs for robots operating in structured environments [2]. However, OLP demands precise calibration and the generated paths do not naturally reflect some relevant parameters associated with manufacturing skills, for example, the orientation and velocity of the robot are defined/estimated offline. Safety issues can be established offline while designing the robotic system [3]. Some studies approach robot programming/teaching by demonstration without physical contact with the robot and with the ability to generalize scenarios from demonstrations [4]. Human pose tracking is one of the main inputs to programming by demonstration systems, using captured data from single or multiple sensors, and delivering tracking data even in the presence of incomplete or corrupted data [5-6]. Vision and laser scanners have been combined to classify parts to be spray coated by a robot [7]. Recently, a vision-based 6D MIMIC solution was proposed for the same process, capturing human skills while applying the coating [8].

The process of adhesive application can be automated by using automatic dispensing machines or robotic arms, usually manually programmed or using OLP. A recent study proposed robot hand-guiding to demonstrate industrial gluing tasks to the robotic system, teaching it [9].

This paper presents an intuitive robot programming framework to capture human manufacturing skills and transform them into robot programs, Fig. 1. This framework is tailored to be used in the process of glass adhesive application, but it can be applied in any other process featuring the capture of human skills from manufacturing tasks such as welding, painting, etc. Experiments on the process of glass adhesive application demonstrated the intuitiveness to use and effectiveness of the system.

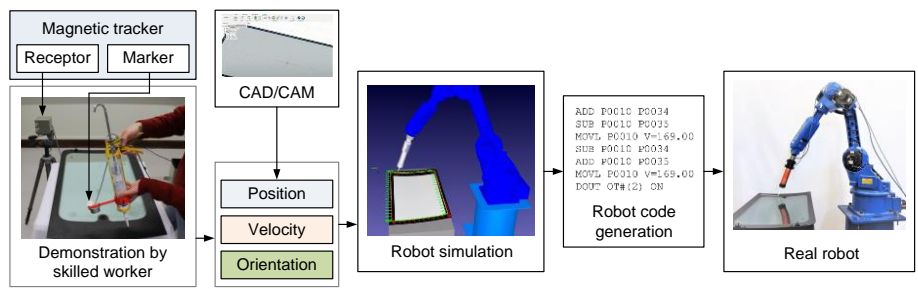

**Fig. 1.** The architecture of the proposed framework for intuitive robot programming by capturing human manufacturing skills.

## 2   Architecture

Fig. 1 shows the proposed framework highlighting the entire process flow, from the demonstration to the real robot in operation. Demonstrations from human skilled workers are recorded using a magnetic tracking device attached to the worker tool (manual glue gun). Collected data include the orientations and velocity of the working paths. Positional data are extracted from CAD/CAM since its error when captured by the magnetic tracker, is significant. Since we are dealing with different reference frames, paths data are transformed in Cartesian space and validated in a simulation environment following an iterative process. Robot programs are generated and transferred to the real robot, which successfully mimics the demonstrated manufacturing skills in the process of glass adhesive application.

### 2.1   Capturing Human Skills

The marker of the magnetic tracking system (Polhemus LIBERTY) is attached to the glue gun while the human demonstrates the process of glass adhesive application, capturing the operator's expertise.

The magnetic tracker provides 6 DOF pose data in relation to the receptor, i.e., positional data in space ($x$, $y$, $z$) and orientation in the form of Euler angles (azimuth/yaw $\psi$, elevation/pitch $\theta$, roll $\phi$). Owing to its electromagnetic nature, when the tracker is nearby metallic/magnetic surfaces or the distance between the marker and the receptor is relatively high, the accuracy is compromised [10]. In such a context, we conducted several experimental tests to evaluate how those elements compromise the accuracy of the system, analyzing the magnitude of the error and its behavior. Fig. 2 shows pose data captured by the magnetic tracker following a rectangular geometry in the xy plane (coordinates along z-axis should be constant). There is a relatively small positional error in measured x and y positional data. This error is not only due to the factors mentioned above (magnetic disturbances and receptor-marker distance) but also due to the human error when manually following the straight lines of the rectangle. On the other hand, the error on positional z coordinates is enormous. As can be seen in Fig. 2, the error increases as the relative receptor-marker distance increases, reaching 60 millimeters. This error makes positional data unusable in our framework. It can also be concluded that the sensor's accuracy is not significantly affected by the presence of metallic objects. Concerning the orientation, the values reported by the magnetic tracker indicate that orientation error is residual, so the orientation data can be used.

Since we have the CAD drawings of the glasses, the path positional data can be directly extracted from CAD/CAM. The orientation can also be extracted from CAD/CAM, for example from the normal to the glass surface, but they do not represent the human skills in what concerns the orientation of the glue gun along the path. Fig. 3 shows the combination of positional data from CAD/CAM with orientation data from the magnetic tracker representing human skills. These resulting pose data are post-processed to reduce the sample size and match those points to the CAD/CAM positional path.

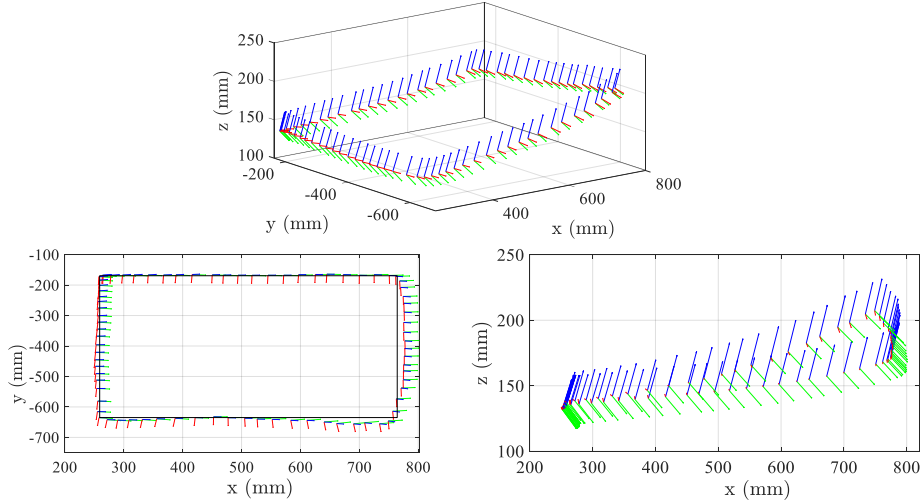

**Fig. 2.** Captured poses from the magnetic tracker following a rectangular geometry in xy plane. The positional error is visible, especially along z-axis.

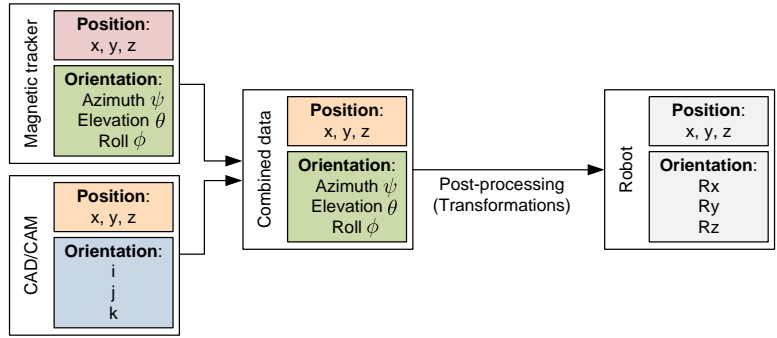

**Fig. 3.** Positional data extracted from CAD/CAM combined with orientation data from the magnetic tracker. The post-processing (homogeneous transformations) output Euler angles following the x-y-z static convention accepted by the robot model we are using.

### 2.2 Coordinated Systems and Transformations

Since we have different reference frames, Fig. 4, collected data from the magnetic tracker and CAD/CAM have to be transformed into a coordinated system that is understandable by the robot. The Euler angles z-y-x from the magnetic tracker represent the current orientation of the marker concerning the reference receptor and can be described by the following rotation matrix:

$$\mathbf{M}_{Z'Y'X'}(\psi,\theta,\phi) = \begin{bmatrix} c\psi c\theta & c\psi s\theta s\phi - s\psi c\phi & c\psi s\theta c\phi + s\psi s\phi \\ s\psi c\theta & -s\psi s\theta s\phi + c\psi c\phi & -s\psi s\theta c\phi - c\psi s\phi \\ -s\theta & c\theta s\phi & c\theta c\phi \end{bmatrix} \quad (1)$$

Where *c* is the cosine and *s* is the sine. The robot we are using, MOTOMAN hp6, requires as input the fixed axis angles x-y-z in which the following matrix is equivalent to the previous one (1):

$$\mathbf{M}_{XYZ}(\phi,\theta,\psi) = \begin{bmatrix} c\psi c\theta & c\psi s\theta s\phi - s\psi c\phi & c\psi s\theta c\phi + s\psi s\phi \\ s\psi c\theta & s\psi s\theta s\phi + c\psi c\phi & s\psi s\theta c\phi - c\psi s\phi \\ -s\theta & c\theta s\phi & c\theta c\phi \end{bmatrix} \quad (2)$$

Fig. 4 shows the reference frames considered for the proposed systems where {*F*} is the floor reference frame, {*S*} is the magnetic tracker's sensor reference frame, {*R*} is the robot reference frame, and {*E*} is the end-effector reference frame. The following transformation applies:

$$^{E}_{S}\mathbf{T} = {}^{E}_{R}\mathbf{T}\,{}^{R}_{F}\mathbf{T}\,{}^{F}_{S}\mathbf{T} \quad (3)$$

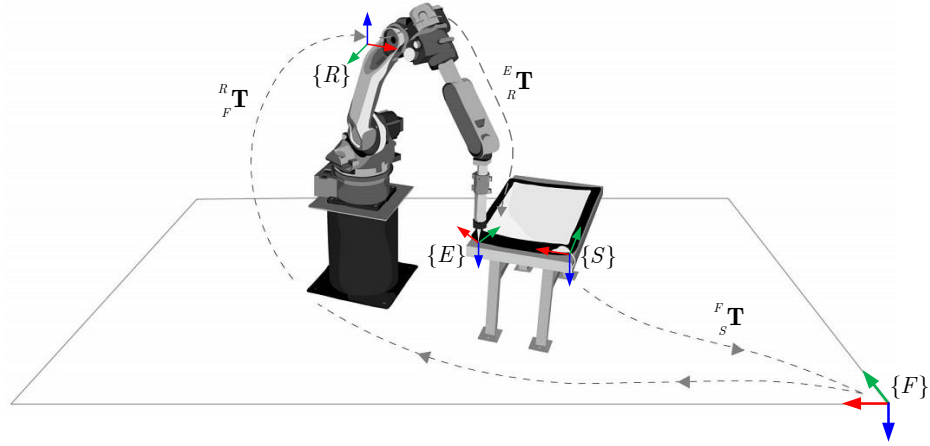

**Fig. 4.** Set of transformations for the proposed framework.

## 3 Experiments and Results

Data collected from human demonstration (orientations) and CAD/CAM (positions), after post-processing, serve as input to the RoboDK robot simulation system. The robot is a Yaskawa MOTOMAN hp6 equipped with the NX100 controller and the glue gun is a Beta 1947H pneumatic bounding gun. The glass is a rear side window from the automotive industry which has a slight curvature, Fig. 5. In the first stage, the robot paths are visualized and analyzed in the RoboDK environment in an iterative process. Fig. 6 shows the paths for the use case of glass adhesive application. The positions are coherent with the glass geometry with no error since the points nominal values are acquired from CAD/CAM. The captured orientations present an error estimated in the range 1º to 5º. This error does not affect the performance of the process to capture the human skills for the glass adhesive application.

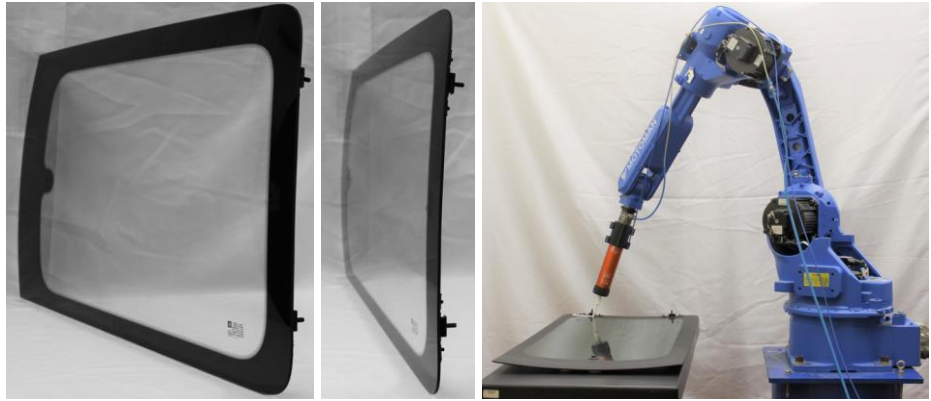

**Fig. 5.** Automotive glass where it is visible the curvature (left and middle) and the robotic cell (right).

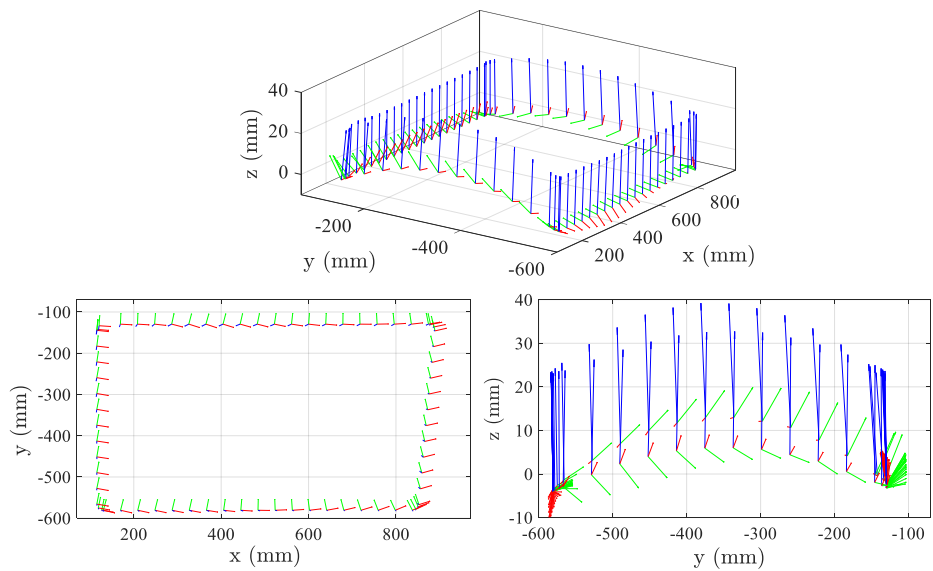

**Fig. 6.** Simulation of the robot paths for the glue application on the automotive glass.

The experimental tests demonstrated the efficiency of the proposed system in capturing human skills and transforming them into robot programs. The human skilled worker just needs to demonstrate the task a single time for each glass model. The generated working paths are validated and adjusted in a simulation environment before transferring them to the real robot. Analyzing the experimental tests from a practical perspective it can be stated that the demonstrated task of glass adhesive application was realized with success by the robot, Fig. 7. The proposed industrial use case demonstrated the feasibility of the proposed method, as demonstrated in auxiliary multimedia materials [11]. Compared to the traditional method used to program robots for this process, using manual teach-in, this framework helps to capture the human skills so that the efficiency of the task is significantly improved.

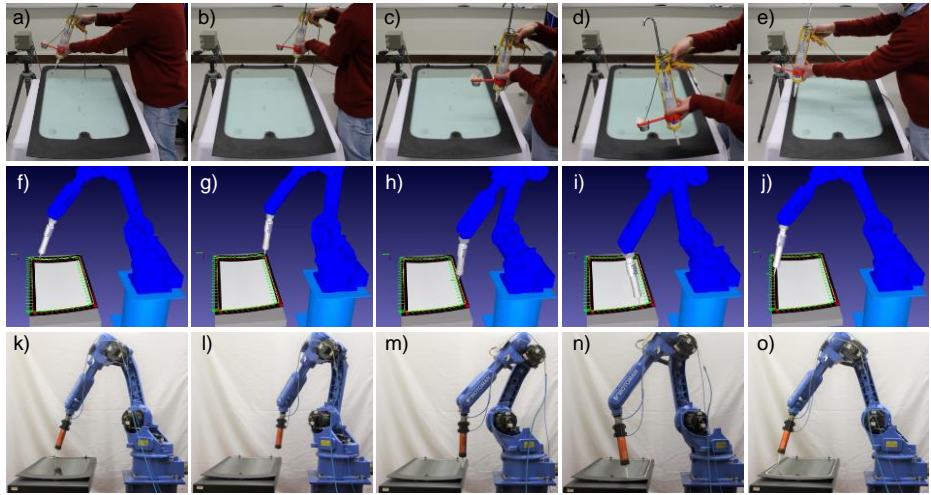

**Fig. 7.** A skilled operator demonstrates the task of glass adhesive application a-e), captured paths data are simulated in OLP allowing to adjust robot poses and anticipate unexpected actions f-j) and the real robot performs the demonstrated tasks k-o).

## 4    Conclusion and Future Work

This paper presented an innovative framework addressing intuitive robot programming by capturing human manufacturing skills from the process of glass adhesive application. It demonstrated effectiveness in generating robot programs from human demonstrations combined with CAD/CAM, making robot programming intuitive and available to non-experts. This means that no advanced skills in robotics are needed to use such functionalities, only basic knowledge in robotics and CAD/CAM are necessary. The implemented industrial use case successfully demonstrated the feasibility of the proposed method, where in few minutes we have the real robot running the demonstrated task, improving quality, efficiency and productivity.

Future work will be dedicated to automating the calibration process and making the framework more generalist, flexible, and easier to use in other manufacturing processes featuring labor skills capturing. The framework will be validated and customized by different people, including industrial workers, targeting its industrial application.

**Acknowledgments.** This research was funded by the European Community's HORIZON 2020 programme under grant agreement No. 958303 (PENELOPE) and by Fundação para a Ciência e a Tecnologia UIDB/00285/2020.

# References

1. Rodriguez-Guerra, D., Sorrosal, G., Cabanes, I., Calleja, C.: Human-Robot Interaction Review: Challenges and Solutions for Modern Industrial Environments. IEEE Access. 9, 108557–108578 (2021). https://doi.org/10.1109/ACCESS.2021.3099287.
2. Zheng, C., An, Y., Wang, Z., Wu, H., Qin, X., Eynard, B., Zhang, Y.: Hybrid offline programming method for robotic welding systems. Robot. Comput. Integr. Manuf. 73, 102238 (2022). https://doi.org/10.1016/j.rcim.2021.102238.
3. Saenz, J., Behrens, R., Schulenburg, E., Petersen, H., Gibaru, O., Neto, P., Elkmann, N.: Methods for considering safety in design of robotics applications featuring human-robot collaboration. Int. J. Adv. Manuf. Technol. 107, 2313–2331 (2020). https://doi.org/10.1007/s00170-020-05076-5.
4. Wang, Y., Jiao, Y., Xiong, R., Yu, H., Zhang, J., Liu, Y.: MASD: A Multimodal Assembly Skill Decoding System for Robot Programming by Demonstration. IEEE Trans. Autom. Sci. Eng. 15, 1722–1734 (2018). https://doi.org/10.1109/TASE.2017.2783342.
5. Simao, M.A., Gibaru, O., Neto, P.: Online Recognition of Incomplete Gesture Data to Interface Collaborative Robots. IEEE Trans. Ind. Electron. 66, 9372–9382 (2019). https://doi.org/10.1109/TIE.2019.2891449.
6. Amorim, A., Guimares, D., Mendona, T., Neto, P., Costa, P., Moreira, A.P.: Robust human position estimation in cooperative robotic cells. Robot. Comput. Integr. Manuf. 67, 102035 (2021). https://doi.org/10.1016/j.rcim.2020.102035.
7. Ferreira, M., Moreira, A.P., Neto, P.: A low-cost laser scanning solution for flexible robotic cells: spray coating. Int. J. Adv. Manuf. Technol. 58, 1031–1041 (2012). https://doi.org/10.1007/s00170-011-3452-x.
8. Arrais, R., Costa, C.M., Ribeiro, P., Rocha, L.F., Silva, M., Veiga, G.: On the development of a collaborative robotic system for industrial coating cells. Int. J. Adv. Manuf. Technol. 115, 853–871 (2021). https://doi.org/10.1007/s00170-020-06167-z.
9. Iturrate, I., Kramberger, A., Sloth, C.: Quick Setup of Force-Controlled Industrial Gluing Tasks Using Learning From Demonstration. Front. Robot. AI. 8, 1–18 (2021). https://doi.org/10.3389/frobt.2021.767878.
10. Tuffaha, M., Stavdahl, O., Stensdotter, A.-K.: Modeling Movement-Induced Errors in AC Electromagnetic Trackers. IEEE Trans. Vis. Comput. Graph. 28, 1597–1607 (2022). https://doi.org/10.1109/TVCG.2020.3019700.
11. Demonstration video, https://youtu.be/CXpUAHYogaw, last accessed 2022/04/26.